\documentclass[letterpaper]{article}
\usepackage{aaai}
\usepackage{times}
\usepackage{helvet}
\usepackage{courier}
\usepackage{natbib}
\usepackage{url}
\usepackage{array}
\usepackage{booktabs}
\usepackage{amsmath}
\usepackage{amssymb}
\usepackage{mathtools}
\usepackage{multirow}
\usepackage{graphicx}
\usepackage{subcaption}
\usepackage{makecell}

\usepackage{hyperref}

\usepackage{etoolbox}
\patchcmd{\maketitle}{\@copyrightspace}{}{}{}

\newcommand{\E}{\mathbb{E}}

\newcommand{\R}{\mathbb{R}}
\newcommand{\N}{\mathcal{N}}

\begin{document}
%
\makeatletter
\gdef\copyright@on{F}
\makeatother
\setcounter{secnumdepth}{2}
\title{VSPrefill: Vertical-Slash Sparse Attention with Lightweight Indexing for Long-Context Prefilling}

\author{Guanzhong Chen \\ School of Computer Science and Technology \\ Tongji University
}
\maketitle

\pagestyle{plain}
\thispagestyle{plain}

\begin{abstract}
\begin{quote}

The quadratic complexity of self-attention during the prefill phase impedes long-context inference in large language models. Existing sparse attention methods face a trade-off among context adaptivity, sampling overhead, and fine-tuning costs. We propose VSPrefill, a mechanism requiring lightweight training that uses the vertical-slash structural pattern in attention distributions. Our compact VSIndexer module predicts context-aware importance scores for vertical columns and slash diagonals from key-value representations augmented with RoPE. This approach constructs sparse masks with linear complexity without modifying the backbone parameters. During inference, an adaptive cumulative-threshold strategy allocates sparsity budgets per layer, while a fused kernel executes attention with on-the-fly index merging. Evaluated on Qwen3-4B-Instruct and LLaMA-3.1-8B-Instruct across the LongBench and RULER benchmarks, VSPrefill preserves 98.35\% of the full attention accuracy while delivering a 4.95$\times$ average speedup at a context length of 128k. These results establish a new Pareto frontier in the trade-off between accuracy and efficiency.

\end{quote}
\end{abstract}

\begin{figure*}[t]
    \centering
    \includegraphics[width=\linewidth]{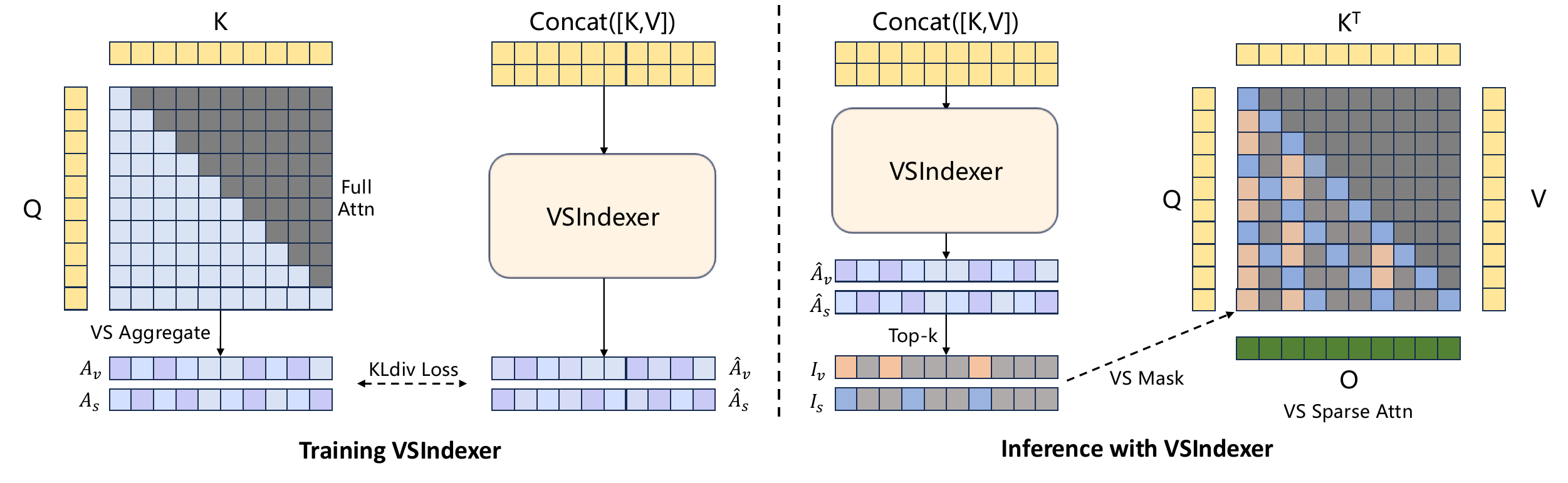}
    \caption{Overview of VSPrefill. The VSIndexer employs a shared-weight bilayer linear network that accepts concatenated key-value pairs as input and outputs vertical and slash attention scores, denoted as $\hat{A}_v$ and $\hat{A}_s$. These scores are trained to approximate the ground-truth full attention weights aggregated along the corresponding directional patterns. During inference, $\hat{A}_v$ and $\hat{A}_s$ undergo top-k selection with a dynamic sparsity budget to construct the vertical-slash sparse attention mask. Implementation details, including the architecture of VSIndexer, RoPE, and sparse attention computation, are omitted for clarity.}
    \label{fig:wide}
\end{figure*}

\section{Introduction}\label{sec:intro}

Large language models (LLMs) have transformed paradigms across science and industry by demonstrating remarkable capabilities in comprehending and generating natural language. To address complex demands such as ultra-long document analysis and repository-level code generation, leading models like Qwen \citep{yang2025qwen3}, Gemini \citep{team2023gemini}, and Claude \citep{caruccio2024claude} have scaled their context windows to the million-token level.

However, this expansion introduces a computational bottleneck due to the quadratic complexity of the self-attention mechanism \citep{fu2024challenges}. As sequence length increases, the Time-to-First-Token (TTFT) grows significantly, which degrades interactivity and increases deployment costs. This challenge has motivated research into sparse attention mechanisms, based on the insight that full attention matrices are intrinsically sparse \citep{zheng2024nsa,li2024snapkv}.

Existing solutions generally fall into two categories, yet each faces limitations. Static approaches, such as BigBird \citep{zaheer2020bigbird} and StreamingLLM \citep{xiao2023streamingllm}, employ fixed, context-agnostic patterns. While efficient, their rigidity often fails to capture input-specific dependencies, leading to accuracy degradation. Dynamic approaches attempt to resolve this by adapting to the context. Training-free methods like Minference \citep{minference2024} and FlexPrefill \citep{lai2025flexprefill} estimate patterns on-the-fly, but they often incur high runtime overhead due to iterative sampling. Among trainable methods, NativeSparseAttention \citep{zheng2024nsa} requires fine-tuning the entire backbone, resulting in high training costs. While SeerAttention \citep{seerattn2024} adopts a frozen-backbone strategy, it remains restricted by the quadratic complexity of its 2D block-wise prediction, which limits acceleration potential.

To address these limitations, we introduce VSPrefill, a sparse prefilling mechanism that achieves the accuracy of trainable methods with the efficiency of static patterns. Our approach is grounded in the empirical observation that salient attention weights naturally organize into a vertical-slash structure, which comprises global "heavy hitters" (vertical) and relative position-dependent correlations (slash).

VSPrefill exploits this structural prior through a lightweight, frozen-backbone training paradigm consisting of three main components. First, we design a parameter-efficient VSIndexer that predicts vertical and slash importance scores directly from RoPE-augmented keys and values, decoupling mask construction from the quadratic attention map. Second, we employ a distillation scheme with a customized kernel that aggregates ground-truth vertical-slash distributions, enabling the indexer to learn accurate global patterns without materializing full attention matrices. Finally, we implement an adaptive inference pipeline with a fused TileLang kernel that performs on-the-fly index merging and execution, achieving linear-complexity index prediction.

We evaluate VSPrefill on state-of-the-art LLMs, including Qwen3-4B-Instruct and LLaMA-3.1-8B-Instruct, across the LongBench \citep{bai2024longbench} and RULER \citep{ruler2024} benchmarks. Experimental results demonstrate that VSPrefill significantly accelerates inference speed while preserving, and in some cases enhancing, model fidelity across diverse tasks and sequence lengths.

\section{Background}
\label{sec:background}

\subsection{LLM Inference}

Modern large language models (LLMs) are predominantly
built upon the Transformer architecture (Vaswani et al.,
2017), which stacks multiple decoder layers, each compris-
ing a self-attention module and a multi-layer perceptron
(MLP).

Inference in LLMs operates through two distinct computational phases: the \textit{prefill} stage and the \textit{decoding} stage~\citep{fu2024challenges, zhen-etal-2025-taming}. The prefill stage processes the entire input sequence of length $n$ in a single parallel forward pass. Here, the self-attention mechanism computes scores across all token pairs, which results in a quadratic time complexity $\Theta(n^2)$. Unlike the decoding phase that benefits from linear complexity $\Theta(n)$, the prefill stage becomes computationally prohibitive for long sequences.

Experimental results with Qwen3-4B-Instruct~\citep{yang2025qwen3} on H20 GPUs quantify this bottleneck. For a context of 256k tokens, self-attention consumes 57.76 seconds, representing 89.51\% of the Time-to-First-Token (TTFT). Scaling to 1M tokens dramatically increases this cost, where attention computation exceeds one hour and accounts for 98.56\% of the total latency. These observations confirm that the quadratic complexity of prefill attention constitutes the primary performance barrier in long-context scenarios.

\subsection{Attention Computation}

In the standard attention mechanism, the input sequence is projected into query, key, and value matrices, denoted as $\mathbf{Q}, \mathbf{K}, \mathbf{V} \in \mathbb{R}^{n \times d}$, where $n$ represents the sequence length and $d$ denotes the head dimension. The core operation computes the output $\mathbf{O}$ via the Scaled Dot-Product Attention:
\begin{align}
    \mathbf{P} &= \mathbf{Q}\mathbf{K}^\top \in \mathbb{R}^{n \times n}, \\
    \mathbf{A} &= \mathrm{softmax}\left(\frac{\mathbf{P}}{\sqrt{d}}\right) \in [0,1]^{n \times n}, \\
    \mathbf{O} &= \mathbf{A}\mathbf{V} \in \mathbb{R}^{n \times d}.
\end{align}
Here, $\mathbf{P}$ represents the unnormalized attention scores capturing pairwise token interactions, and $\mathbf{A}$ denotes the normalized attention probability matrix. While FlashAttention~\citep{dao2022flashattention} optimizes memory access through tiling and kernel fusion, it does not reduce the theoretical FLOPs count. The inherent $\Theta(n^2)$ complexity of the matrix multiplication $\mathbf{Q}\mathbf{K}^\top$ remains a critical bottleneck for processing million-token contexts, necessitating the development of efficient approximations.

\subsection{Sparse Attention Formulation}

\begin{figure}[t]
    \centering
    \includegraphics[width=0.9\linewidth]{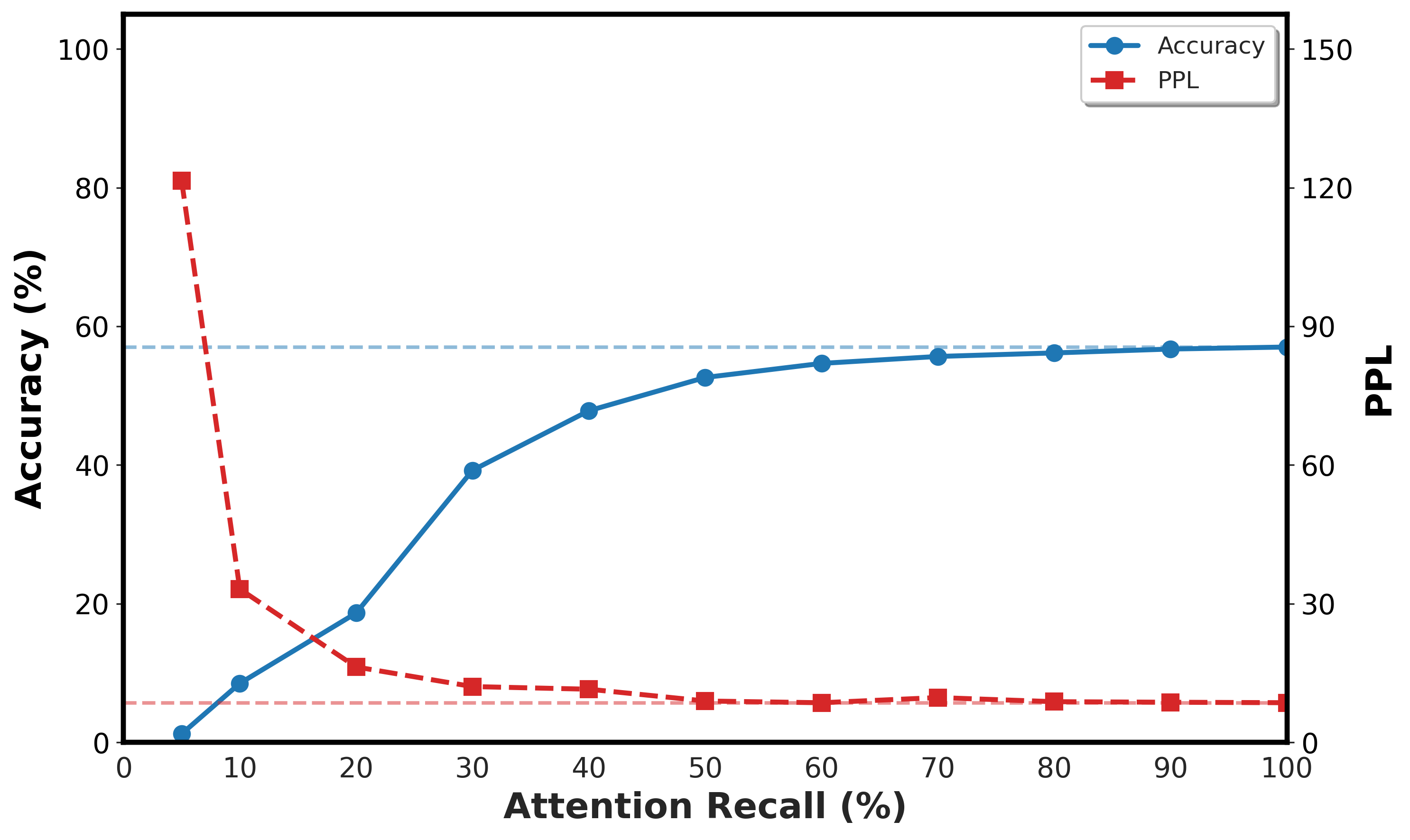}
    \caption{Accuracy and Perplexity Trends Across Different Attention Recall Levels on HotPotQA dataset.}
    \label{fig:accuracy_vs_recall}
\end{figure}

Empirical studies demonstrate that the attention matrix $\mathbf{A}$ is inherently sparse, characterized by a heavy-tailed distribution where a small subset of elements contributes the majority of the probability mass~\citep{zheng2024nsa,li2024snapkv}. This observation motivates sparse attention, which approximates the full computation by retaining only a selected subset of query-key pairs indexed by a sparsity pattern $S$. The masked attention scores are computed as:
\begin{align}
    \mathbf{M}_S[i,j] &=
    \begin{cases}
        0 & \text{if } (i,j) \in S \\
        -\infty & \text{otherwise}
    \end{cases}\\
    \hat{\mathbf{A}} &= \mathrm{softmax}\left(\frac{\mathbf{P}}{\sqrt{d}} + \mathbf{M}_S\right)
\end{align}
To evaluate the information retention of a sparse pattern, we utilize the \textit{Attention Recall} metric. For a given index set $S$, the recall $R(S)$ is defined as the proportion of the original attention mass preserved within the selected indices:
\begin{align}
    R(S) = \frac{1}{n} \sum_{i=0}^{n-1} \sum_{j=0}^{n-1} \mathbf{A}[i,j] \cdot \mathbb{I}\big((i,j) \in S\big),
\end{align}
where $\mathbb{I}(\cdot)$ denotes the indicator function. The relevance of this metric is substantiated in Figure~\ref{fig:accuracy_vs_recall}, which reveals a strong correlation between Attention Recall and downstream metrics. Notably, the model achieves functional viability when Attention Recall exceeds 50\%, characterized by stabilized perplexity and a plateauing accuracy curve. Furthermore, surpassing 90\% recall yields performance indistinguishable from full attention. This monotonic relationship validates the use of Recall as a faithful surrogate objective for maintaining model fidelity.

Consequently, the design of an optimal sparse attention mechanism can be formulated as a constrained optimization problem that maximizes recall subject to a computational budget $k$:
\begin{align}
\max_{S} \quad & R(S) \\
\textrm{s.t.} \quad & |S| \leq k.
\end{align}
This formulation highlights the fundamental trade-off between approximation fidelity and inference efficiency. Our proposed method, VSPrefill, addresses this optimization by predicting the structure of $S$ through a decomposed vertical-slash representation.

\section{Motivation}
\label{sec:motivation}

Recent scholarship has established that the self-attention mechanism possesses inherent sparsity~\citep{zheng2024nsa,ji-etal-2021-distribution,zaheer2020bigbird,li2024snapkv,minference2024}. However, this sparsity is dynamic, exhibiting a heterogeneous nature that fluctuates across model architectures, input contexts, and individual attention heads. Such dynamism presents substantial hurdles for the design of efficient sparse attention mechanisms. In this section, we first characterize these dynamic behaviors through empirical analysis. Subsequently, we identify and provide a theoretical basis for a prevalent structural phenomenon, termed the \textit{vertical-slash pattern}, which underpins these dynamics. Finally, leveraging this structural insight, we decompose the intractable mask prediction problem into two sub-problems with linear complexity.

\begin{figure*}[t]
    \centering
    \begin{subfigure}[b]{0.16\textwidth}
        \centering
        \includegraphics[width=\textwidth]{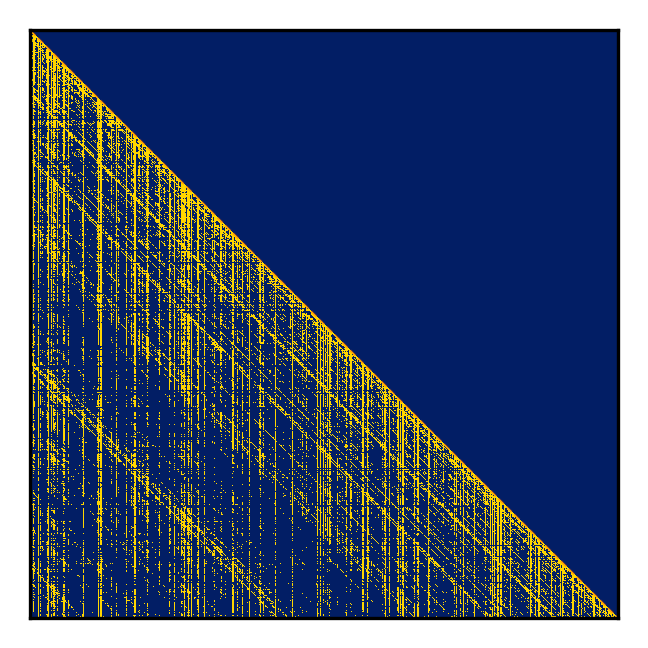}
        \caption{\scriptsize Reference Head}
        \label{fig:viz-base}
    \end{subfigure}
    \hfill
    \begin{subfigure}[b]{0.16\textwidth}
        \centering
        \includegraphics[width=\textwidth]{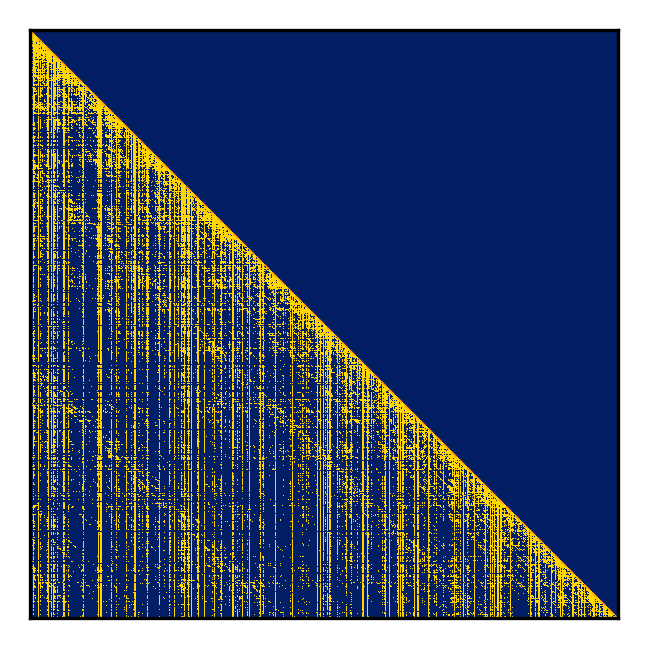}
        \caption{\scriptsize Intra-Group Head}
        \label{fig:viz-intra-group}
    \end{subfigure}
    \hfill
    \begin{subfigure}[b]{0.16\textwidth}
        \centering
        \includegraphics[width=\textwidth]{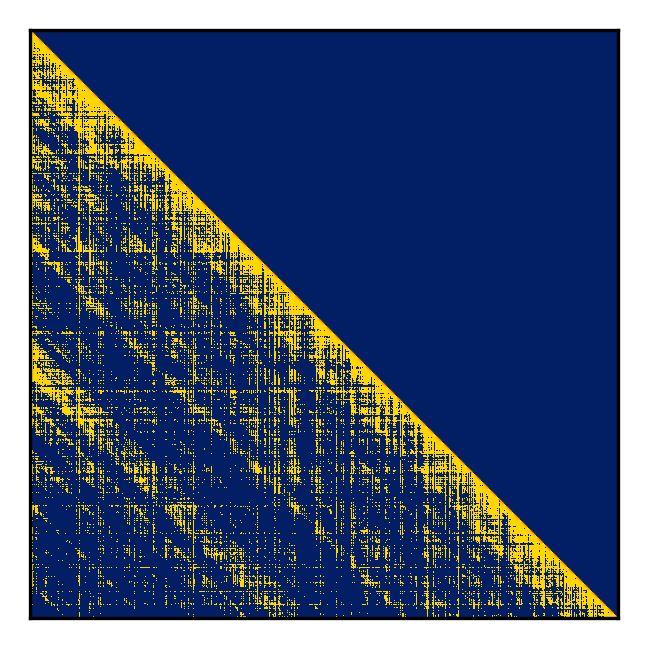}
        \caption{\scriptsize Inter-Group Head}
        \label{fig:viz-inter-group}
    \end{subfigure}
    \hfill
    \begin{subfigure}[b]{0.16\textwidth}
        \centering
        \includegraphics[width=\textwidth]{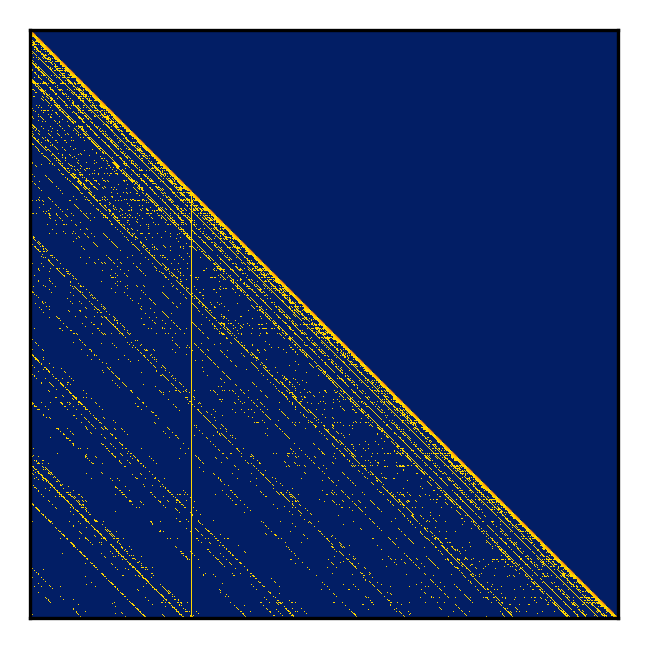}
        \caption{\scriptsize Different Layer}
        \label{fig:viz-deep-layer}
    \end{subfigure}
    \hfill
    \begin{subfigure}[b]{0.16\textwidth}
        \centering
        \includegraphics[width=\textwidth]{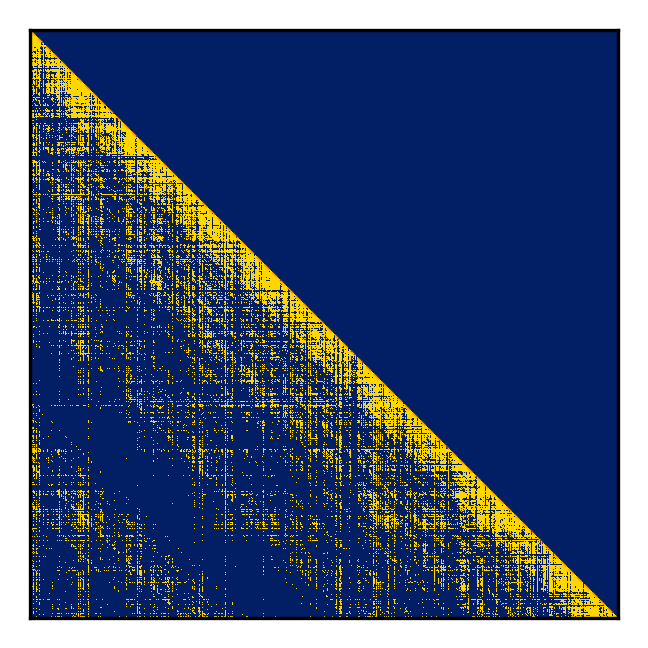}
        \caption{\scriptsize Different Prompt}
        \label{fig:viz-diff-prompt}
    \end{subfigure}
    \hfill
    \begin{subfigure}[b]{0.16\textwidth}
        \centering
        \includegraphics[width=\textwidth]{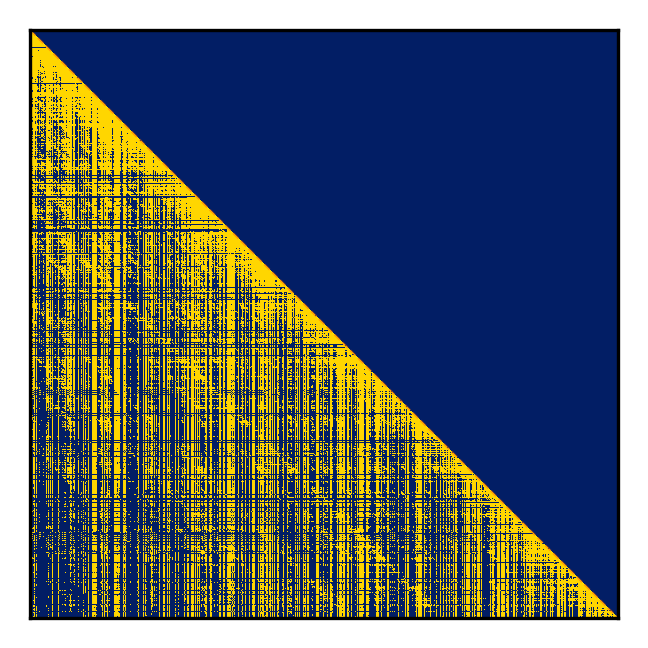}
        \caption{\scriptsize Different Model}
        \label{fig:viz-llama}
    \end{subfigure}
    
    \caption{Visualization of dynamic attention sparsity patterns.
    \textbf{(a-c)} Comparisons reveal high structural similarity within the same KV group (Intra-Group) contrasted with distinct topologies across different groups (Inter-Group).
    \textbf{(d)} Sparsity patterns evolve significantly as network depth increases.
    \textbf{(e-f)} The distribution of salient weights shifts in response to varying input prompts and model architectures.}
    \label{fig:attention_visualize}
\end{figure*}

\subsection{Dynamic Nature of Attention Sparsity}
\label{subsec:dynamic_sparsity}

Through a systematic examination of attention distributions within the Qwen3-4B-Instruct model on the LongBench benchmark~\citep{bai2024longbench}, we demonstrate that attention sparsity is a dynamic property, as visualized in Figure~\ref{fig:attention_visualize}, characterized by the following distinct features:

\begin{itemize}
    \item \textbf{Intra-Group Consistency versus Inter-Group Divergence:} Attention patterns remain highly consistent across heads within the same KV group (Figures~\ref{fig:viz-base}, \ref{fig:viz-intra-group}), validating our strategy of sharing sparse masks per group. Conversely, significant structural divergence exists between distinct KV groups and layers (Figures~\ref{fig:viz-base}, \ref{fig:viz-inter-group}, \ref{fig:viz-deep-layer}), which necessitates group-specific prediction rather than a global uniform pattern.
    
    \item \textbf{Context Sensitivity:} The sparse topology is strictly input-dependent. As shown in Figures~\ref{fig:viz-base} and~\ref{fig:viz-diff-prompt}, varying input prompts induce observably different attention maps even under identical model configurations.
    
    \item \textbf{Model Dependence:} Different architectures yield unique attention signatures, as evidenced by the distinct sparsity profiles of Qwen3-4B-Instruct versus LLaMA-3.1-8B-Instruct (Figures~\ref{fig:viz-base}, \ref{fig:viz-llama}).
\end{itemize}

These dynamic properties render static patterns, such as the fixed structures employed in StreamingLLM~\citep{xiao2023streamingllm}, fundamentally insufficient for capturing ground-truth attention. To maintain high fidelity without performance degradation, a mechanism must dynamically determine both the \textit{topology}, which dictates which tokens to select, and the \textit{cardinality}, which governs the size of the sparse index set.

\subsection{Vertical-Slash Attention Sparse Pattern}
\label{subsec:vertical_slash_pattern}

Our empirical investigation reveals a pervasive and structured motif within the attention matrices of long-context LLMs, which we term the \textit{vertical-slash pattern} (visualized in Figure~\ref{fig:attention_visualize}). This pattern exhibits a dual characteristic. The \textit{vertical} lines represent "heavy hitters," which are global anchor tokens that sustain high attention regardless of distance. Conversely, the \textit{slash} lines denote position-dependent correlations where attention is maintained at specific relative offsets. This structure generalizes the static "A-shaped" prior~\citep{xiao2023streamingllm} by capturing not just local context and initial tokens, but also long-range, periodic dependencies inherent to complex reasoning.

\begin{figure}[t]
    \centering
    \includegraphics[width=0.5\textwidth]{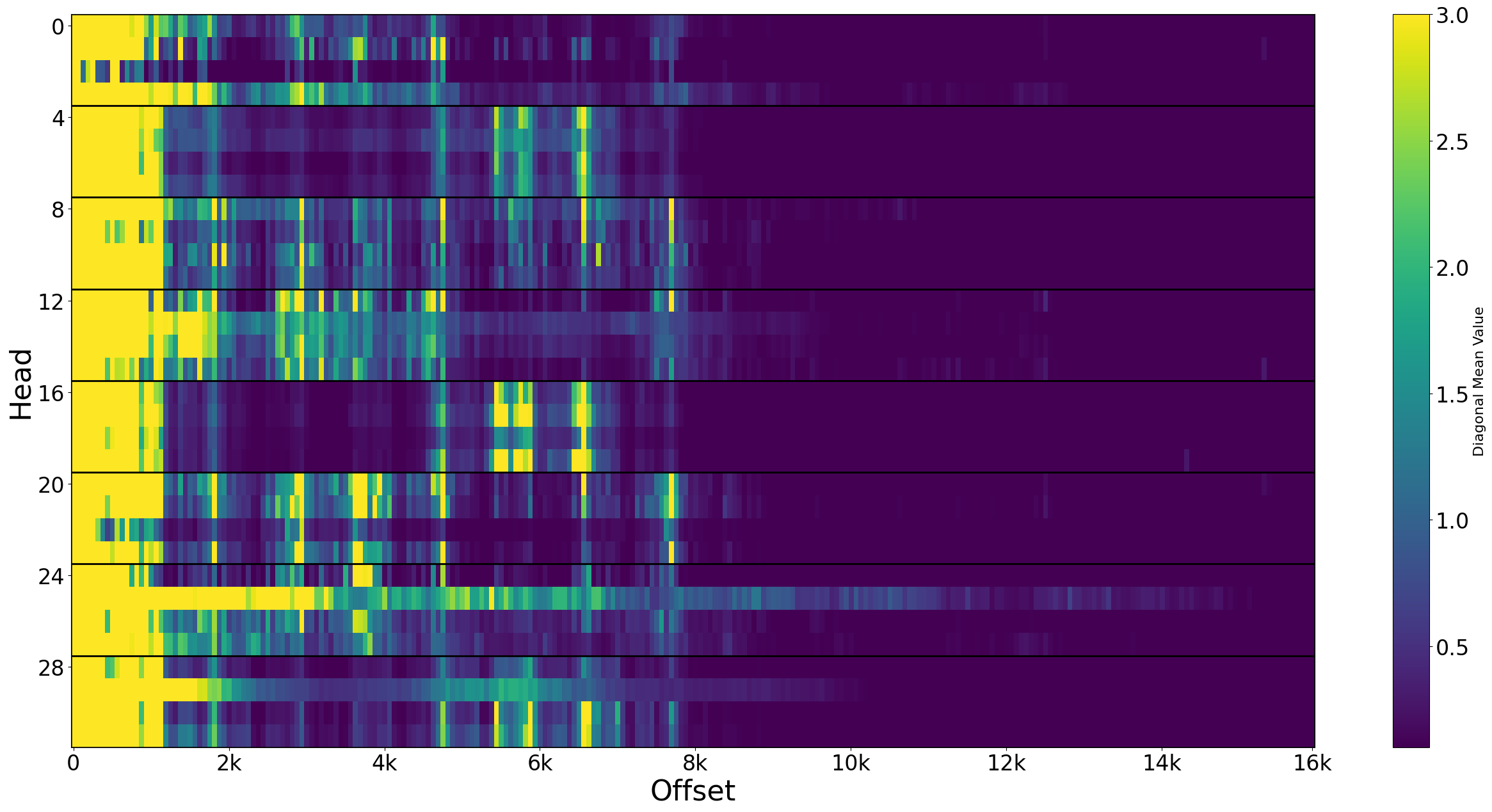}
    \caption{Diagonal-aggregated attention heatmap for Layer 0 of Qwen3-4B-Instruct on LongBench. The x-axis denotes the relative diagonal offset ($i-j$) where 0 corresponds to the main diagonal. The emergence of distinct vertical bands at distal offsets validates the existence of consistent \textit{slash} patterns, characterized by strong correlations at fixed relative distances across multiple heads.}
    \label{fig:attention_aggregated_slash_wo_mean}
\end{figure}

To analyze the genesis of the \textit{slash} component, we visualize the aggregated attention scores along diagonal offsets in Figure~\ref{fig:attention_aggregated_slash_wo_mean}. This heatmap exposes two properties. First, a high-intensity band near the main diagonal validates the strong inductive bias of local sliding windows for maintaining linguistic continuity. Second, discrete, high-activation vertical bands emerge at specific long-range offsets. These bands correspond to the "slash" lines in the full matrix and persist across multiple heads within the same KV group. This observation suggests that sparse attention correlations are systematically established at specific relative distances rather than being randomly dispersed.

We theoretically attribute this phenomenon to the Rotary Positional Embedding (RoPE) mechanism. In Appendix~\ref{sec:slash_pattern_analysis}, we prove that under multivariate Gaussian assumptions for query and key distributions, the attention score expectation evolves as a function of the relative position $i-j$. This derivation mathematically elucidates why attention peaks manifest at regular intervals, providing a solid theoretical foundation for the observed slash pattern.

\subsection{Linear-Complexity Mask Construction}
\label{subsec:problem_decomposition}

The central challenge in sparse attention lies in accurately identifying the most salient query-key pairs within the attention matrix. Existing block-wise prediction methods, such as SeerAttention~\citep{seerattn2024}, fail to escape quadratic complexity. By using the vertical-slash structure, we decompose the mask construction problem from a quadratic search space into two decoupled linear subspaces, thereby reducing the computational complexity to $O(n)$.

Formally, we define the sparse mask $M_S$ as follows:
\begin{align}
M_S[i,j] = 
\begin{cases} 
0 & \text{if } j \in I_v \text{ or } (i - j) \in I_s, \\
-\infty & \text{otherwise},
\end{cases}
\end{align}
where $I_v$ and $I_s$ denote the sets of selected indices for the vertical and slash components, respectively.
This decomposition allows us to reframe the original problem into two independent constrained optimization tasks. For each direction $d \in \{v, s\}$, we formulate the objective as maximizing the Attention Recall ($R$), defined as the proportion of attention mass preserved by the selected indices, subject to a sparsity budget $k_d$. The optimization problem is expressed as:
\begin{equation}
\begin{aligned}
\max_{I_d} \quad & R(I_d) \\
\textrm{s.t.} \quad & |I_d| \leq k_d.
\end{aligned}
\end{equation}
This formulation explicitly addresses the trade-off between approximation fidelity (maximizing $R$) and computational efficiency (adhering to budget $k_d$). Our proposed method, VSPrefill, solves these linear-complexity problems via lightweight training to predict accurate, context-aware sparse patterns with minimal overhead.

\section{Method}
\label{sec:method}

This section presents VSPrefill, a lightweight-training sparse attention mechanism that predicts context-aware vertical-slash patterns with favorable computational efficiency. Although existing approaches such as FlexPrefill~\citep{lai2025flexprefill} and Sample Attention~\citep{zhu2025sampleattention} implicitly exploit vertical-slash structures through query sampling, they face a trade-off: single-point sampling incurs modest overhead yet fails to capture global patterns, whereas multi-point sampling reduces estimation variance at the cost of substantial computation that undermines acceleration benefits. To resolve this, we propose a lightweight trainable module that accurately predicts vertical-slash patterns with minimal overhead. Our method comprises three core components: the VSIndexer prediction module design, vertical-slash aggregation for distillation training, and efficient vertical-slash sparse attention inference.

\subsection{VSIndexer Design}
\label{subsec:vsindexer_design}

The VSIndexer module predicts vertical and slash patterns with minimal additional computation. Its input is formed by concatenating the key matrix $K$ (with Rotary Positional Embedding applied) and the value matrix $V$, forming $X = \mathrm{concat}(K, V) \in \mathbb{R}^{n \times 2d}$. This design uses complementary signal sources: keys augmented with RoPE encode rich positional relationships that substantially contribute to slash pattern detection, while values carry semantic information critical for identifying high-importance vertical columns.

To maximize parameter efficiency and minimize both training and inference overhead, both vertical and slash scoring pathways share an initial linear up-projection layer followed by a non-linear activation $\sigma$:
\begin{align}
    \textbf{X} &= \mathrm{concat}(\textbf{K}, \textbf{V}), \\
    \textbf{Z} &= \sigma(\textbf{X}\textbf{W}_{U} + \textbf{b}_{U}), \\
    \hat A_v &= \mathrm{softmax}(\textbf{Z}\textbf{W}_{V} + \textbf{b}_{V}), \\
    \hat A_s &= \mathrm{softmax}(\textbf{Z}\textbf{W}_{S} + \textbf{b}_{S}),
\end{align}
where $\textbf{W}_U \in \mathbb{R}^{2d \times d_{h}}$ and $\textbf{b}_U \in \mathbb{R}^{d_{h}}$ denote the shared projection parameters. Parameters $\textbf{W}_{V}$, $\textbf{W}_{S}$ and their corresponding biases $\textbf{b}_{V}$, $\textbf{b}_{S}$ constitute independent down-projection layers that generate directional scoring vectors $A_v, A_s \in \mathbb{R}^{n}$, where each element corresponds to the importance score of a vertical column or slash diagonal. The use of softmax mimics the sharp, peak-dominated distributions observed in ground-truth aggregated attention maps, thereby facilitating effective index selection by emphasizing the most salient positions. This design remains parameter-efficient while incurring a linear computational overhead of $\mathcal{O}(nd_{h})$ per attention head.

\subsection{Vertical-Slash Aggregation for Distillation Training}
\label{subsec:distillation_training}

Direct co-training of the VSIndexer with the backbone model proves computationally prohibitive. Instead, we adopt a distillation paradigm that freezes all original parameters of the model and trains only the VSIndexer module using ground-truth attention distributions derived from full attention computation.

\paragraph{Obtaining Ground Truth }
We aggregate full attention weights $\textbf{A} = \mathrm{softmax}(\textbf{QK}^\top / \sqrt{d})$ along vertical columns and slash diagonals to obtain target distributions $A_{v}$ and $A_{s}$. Conceptually, higher aggregated scores indicate greater contributions to attention recall, making accurate prediction of these distributions essential for effective sparse pattern selection. However, explicit extraction of $A$ is infeasible for long sequences because modern LLM inference relies on FlashAttention~\citep{dao2022flashattention}, which fuses all attention operations into a single kernel and deliberately avoids materializing intermediate attention matrices to maintain favorable memory complexity.

To address this challenge, we implement a customized FlashAttention kernel using TileLang that preserves the original computation flow while performing online aggregation during block-wise attention computation. Specifically, after computing partial attention scores within each tile, our kernel accumulates contributions along vertical columns and slash diagonals without storing the full $n \times n$ matrix. Since aggregation compresses the $n \times n$ attention matrix into two $n$-dimensional vectors whose elements sum to $n$, we normalize both $A_{v}$ and $A_{s}$ by dividing by $n$ to form proper probability distributions suitable for distillation.

\paragraph{Loss function design }
We employ KL divergence to align the predicted distributions $\hat{A}_{v}$, $\hat{A}_{s}$ with their ground-truth counterparts $A_{v}$, $A_{s}$:
\begin{align}
    A_{v}, A_{s} &= \mathrm{VSAggregate}\bigl(\mathrm{softmax}(\frac{\textbf{Q}\textbf{K}^\top } {\sqrt{d}})\bigr), \\
    \hat{A}_{v}, \hat{A}_{s} &= \mathrm{VSIndexer}(\textbf{K}, \textbf{V}), \\
    \mathcal{L} &= D_{\mathrm{KL}}(\hat{A}_{v} \,\|\, A_{v}) + D_{\mathrm{KL}}(\hat{A}_{s} \,\|\, A_{s}).
\end{align}
KL divergence encourages distributional matching rather than absolute magnitude learning, which proves more effective than MSE for capturing the skewed, peaky nature of attention distributions. Moreover, separating losses for vertical and slash components enables the model to learn distinct characteristics of each pattern independently.

\subsection{Efficient Vertical-Slash Sparse Attention Inference}
\label{subsec:efficient_inference}

After distillation training, the VSIndexer outputs scoring vectors $\hat{A}_{v}$ and $\hat{A}_{s}$ for vertical columns and slash offsets, respectively.

\paragraph{Adaptive Sparse Index Selection}
To achieve optimal resource allocation, we employ a cumulative-threshold strategy that dynamically determines sparsity budgets based on the distribution of predicted importance scores. Unlike static or length-proportional methods, our approach first establishes the sparsity budgets $k_v$ and $k_s$by identifying the minimum top-ranked elements needed to meet the cumulative mass thresholds $\tau_v$ and $\tau_s$. Formally, for each direction $d \in \{v, s\}$, the budget determination and subsequent index selection are defined as:
\begin{align}
    k_{d} &= \min\left\{k \,\middle|\, \sum_{i=1}^{k} \mathrm{sort}(\hat{A}_{d})_{i} \geq \tau_{d}\right\}, \\
    I_{d} &= \mathrm{TopK}(\hat{A}_{d}, k_{d}).
\end{align}
This data-driven mechanism instills adaptivity across three critical dimensions: 1) context-awareness, where budgets automatically expand for high-complexity inputs and contract for simpler ones; 2) layer-specificity, allowing distinct layers to autonomously prioritize global information aggregation (vertical patterns) or local linguistic coherence (slash patterns); and 3) model-dependence, as the learned scoring distributions inherently capture the unique attention signatures of the underlying architecture.

\paragraph{Fused attention kernel implementation}
Implementing vertical-slash sparse attention efficiently presents non-trivial engineering challenges, as standard FlashAttention tiling strategies assume contiguous memory access patterns which are violated by the non-contiguous query-key pairs in our sparsity pattern, and naive index precomputation would incur prohibitive memory overhead. To address these issues, our fused kernel incorporates three key optimizations within a unified pipeline. First, we adopt a tiling strategy that reads sparse indices block-wise and fetches corresponding key-value pairs on demand, thereby reusing the optimized fused computation pipeline of FlashAttention. Second, rather than materializing a full index matrix, we compute merged indices on-the-fly during each query block processing; since both vertical and slash index lists are naturally sorted, their union is generated via an efficient GPU-parallel merge operation based on the Merge Path algorithm~\citep{green2012gpu}, which partitions the workload evenly across threads. Third, the entire pipeline is implemented in TileLang to maximize hardware utilization and minimize kernel launch overhead, maintaining high GPU occupancy while avoiding excessive intermediate storage.


\begin{table*}[t]
    \centering
    \captionsetup{font=small, skip=4pt}
    \renewcommand{\arraystretch}{1.25}
    \setlength{\tabcolsep}{4.5pt}
    \small
    \caption{Performance evaluation on the RULER benchmark across context lengths ranging from 4k to 128k. Bold values indicate the highest score within each model group (excluding FlashAttn); underlined values indicate the second-highest score. Speedup is relative to full attention.}
    \begin{tabular}{@{}
        >{\centering\arraybackslash}p{1.65cm} |
        >{\centering\arraybackslash}p{1.45cm} |
        *{6}{>{\centering\arraybackslash}p{0.68cm}} |
        *{2}{>{\centering\arraybackslash}p{1.18cm}}
        @{}}
        \toprule
        \multirow{2}{*}{\textbf{Model}} & 
        \multirow{2}{*}{\textbf{Method}}  &
        \multicolumn{6}{c|}{\textbf{Length}} &
        \multirow{2}{*}{\textbf{\makecell{Avg.\\Score}}} & 
        \multirow{2}{*}{\textbf{\makecell{Avg.\\Speedup}}} \\
        & &
        \textbf{4k} & \textbf{8k} & \textbf{16k} & \textbf{32k} & \textbf{64k} & \textbf{128k} \\
        \midrule
        \multirow{5}{*}{\makecell{Qwen3-4B-\\Instruct}} 
        & FlashAttn            & 77.99 & 79.40 & 80.23 & 78.96 & 84.00 & 77.41 & 79.67 & -- \\
        & StrLLM & 70.85 & 64.84 & 58.47 & 52.31 & 47.56 & 36.16 & 55.03 & 4.79 \\
        & FlexPre & 74.32 & 76.62 & 76.87 & 76.87 & 79.44 & \underline{74.44} & \underline{76.43} & 1.83 \\
        & SeerAttn    & \underline{77.10} & \textbf{78.24} & \underline{79.92} & \underline{77.16} & \underline{81.43} & 64.45 & 76.38 & 1.67  \\
        & VSPrefill     & \textbf{77.46} & \underline{77.27} & \textbf{80.00} & \textbf{78.29} & \textbf{83.47} & \textbf{75.15} & \textbf{78.61} & 1.91 \\
        \midrule
        \multirow{5}{*}{\makecell{LLaMA-3.1-\\8B-Instruct}} 
        & FlashAttn & 96.48 & 94.70 & 89.84 & 79.82 & 77.48 & 74.27 & 85.43 & -- \\
        & StrLLM      & 82.02 & 69.76 & 45.56 & 30.77 & 12.52 & 6.64 & 41.21 & 6.71 \\
        & FlexPre       & 93.19 & 93.41 & \textbf{90.83} & 77.58 & \underline{75.27} & \textbf{73.80} & \underline{84.01} & 1.69 \\
        & SeerAttn & \underline{93.83} & \underline{93.83} & 90.00 & \textbf{78.86} & 75.19 & 72.29 & 84.00 & 1.59  \\
        & VSPrefill        & \textbf{96.67} & \textbf{94.50} & \underline{90.68} & \underline{78.26} & \textbf{75.34} & \underline{72.76} & \textbf{84.70} & 1.75 \\
        \bottomrule
    \end{tabular}
    
    \label{tab:attention-comparison}
\end{table*}

\begin{table*}[t]
    \centering
    \captionsetup{font=small, skip=4pt}
    \renewcommand{\arraystretch}{1.25}
    \setlength{\tabcolsep}{4.5pt}
    \caption{Performance evaluation on the LongBench benchmark across 13 diverse tasks. Bold values indicate the highest score within each model group (excluding FlashAttn); underlined values indicate the second-highest score.}
    \small
    
    \begin{tabular}{@{}
        >{\centering\arraybackslash}p{1.65cm} |
        >{\centering\arraybackslash}p{1.45cm} |
        *{13}{>{\centering\arraybackslash}p{0.68cm}}
        >{\centering\arraybackslash}p{0.85cm} 
        @{}}
        \toprule
        \textbf{Model} & \textbf{Method} &
        \textbf{\rotatebox{45}{Qasper}} & 
        \textbf{\rotatebox{45}{MNews}} & 
        \textbf{\rotatebox{45}{Trec}} & 
        \textbf{\rotatebox{45}{2WikiQA}} & 
        \textbf{\rotatebox{45}{LCC}} & 
        \textbf{\rotatebox{45}{MF-en}} & 
        \textbf{\rotatebox{45}{GovRep}} & 
        \textbf{\rotatebox{45}{PsgRetr}} & 
        \textbf{\rotatebox{45}{PsgCnt}} & 
        \textbf{\rotatebox{45}{SamSum}} & 
        \textbf{\rotatebox{45}{RBench}} & 
        \textbf{\rotatebox{45}{HotPot}} & 
        \textbf{\rotatebox{45}{TriviQA}} & 
        \textbf{\rotatebox{45}{Avg.}} \\
        \midrule
        \multirow{4}{*}{\makecell{Qwen3-4B-\\Instruct}} 
        & FlashAttn   & 40.66 & 22.12 & 72.67 & 40.28 & 6.41 & 50.53 & 30.75 & 100.00 & 1.45 & 35.98 & 8.25 & 57.61 & 85.49 & 42.48 \\
        & StrLLM      & 31.62 & 22.22 & 73.00 & 26.57 & \textbf{6.07} & 33.76 & 30.60 & 38.17 & \textbf{2.89} & \underline{36.64} & 7.13 & 35.83 & \textbf{84.29} & 32.98 \\
        & FlexPre     & 37.90 & \textbf{22.40} & \textbf{74.00} & 31.50 & 4.68 & \textbf{50.85} & 30.22 & \underline{51.78} & \underline{2.06} & \textbf{36.72} & 7.24 & 47.91 & 83.24 & 36.96 \\
        & SeerAttn    & \underline{41.90} & 22.11 & 70.33 & \textbf{38.68} & 5.00 & 47.31 & \underline{30.64} & \textbf{100.00} & 1.48 & 34.08 & \underline{7.46} & \textbf{57.39} & 84.04 & \underline{41.57} \\
        & VSPrefill   & \textbf{42.39} & \underline{22.33} & \underline{73.67} & \underline{37.50} & \underline{5.09} & \underline{48.11} & \textbf{30.71} & \textbf{100.00} & 1.21 & 36.15 & \textbf{9.24} & \underline{52.59} & \underline{84.21} & \textbf{41.78} \\
        \midrule
        \multirow{4}{*}{\makecell{LLaMA-3.1-\\8B-Instruct}} 
        & FlashAttn   & 42.98 & 26.18 & 8.00 & 43.46 & 26.28 & 55.25 & 34.93 & 99.67 & 11.72 & 8.13 & 22.81 & 60.94 & 88.76 & 40.70 \\
        & StrLLM      & 34.47 & 25.59 & 5.67 & 29.12 & \underline{26.97} & 32.53 & 34.35 & 69.00 & 10.54 & \underline{8.27} & \underline{23.75} & 36.66 & \textbf{90.49} & 32.88 \\
        & FlexPre     & 39.79 & \underline{25.93} & \textbf{8.67} & 33.15 & \textbf{27.73} & \underline{53.49} & 34.30 & \underline{80.67} & 5.86 & \textbf{16.59} & \textbf{26.22} & 55.31 & \underline{89.58} & 38.25 \\
        & SeerAttn    & \underline{42.74} & 24.88 & \underline{6.33} & \underline{41.04} & 26.12 & 52.76 & \underline{35.10} & \textbf{99.33} & \textbf{11.60} & 8.03 & 22.56 & \underline{57.73} & 87.59 & \underline{39.68} \\
        & VSPrefill   & \textbf{43.52} & \textbf{26.02} & 4.67 & \textbf{41.52} & 26.05 & \textbf{54.04} & \textbf{35.12} & \textbf{99.33} & \underline{11.39} & 8.06 & 22.46 & \textbf{58.58} & 88.41 & \textbf{39.94} \\
        \bottomrule
    \end{tabular}
    
    \label{tab:longbench-result}
\end{table*}

\begin{figure*}[t]
    \centering
    \begin{subfigure}[b]{0.32\textwidth}
        \centering
        \includegraphics[width=\textwidth]{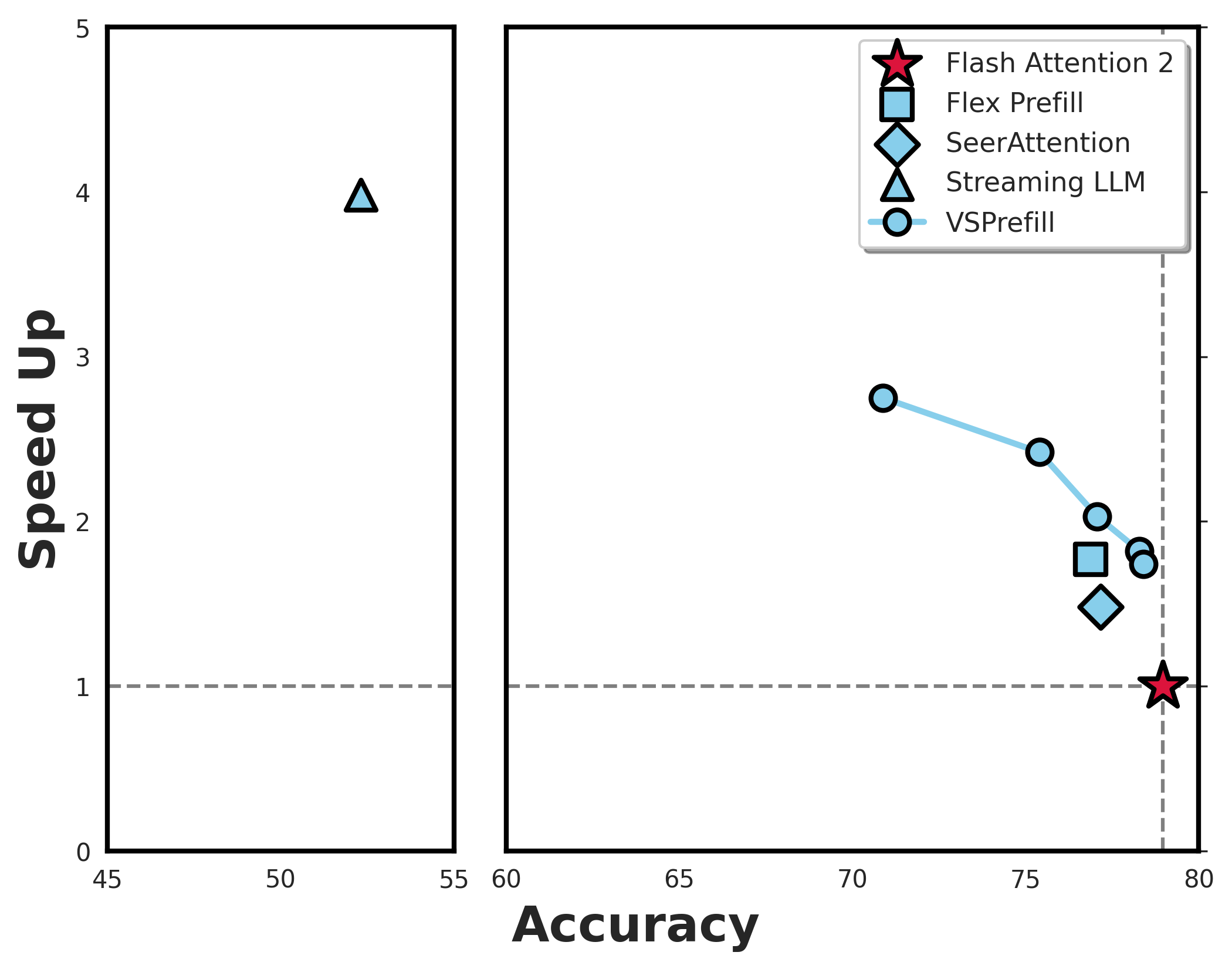}
        \caption{32K Context}
        \label{fig:accuracy_vs_speedup_32k}
    \end{subfigure}
    \hfill
    \begin{subfigure}[b]{0.32\textwidth}
        \centering
        \includegraphics[width=\textwidth]{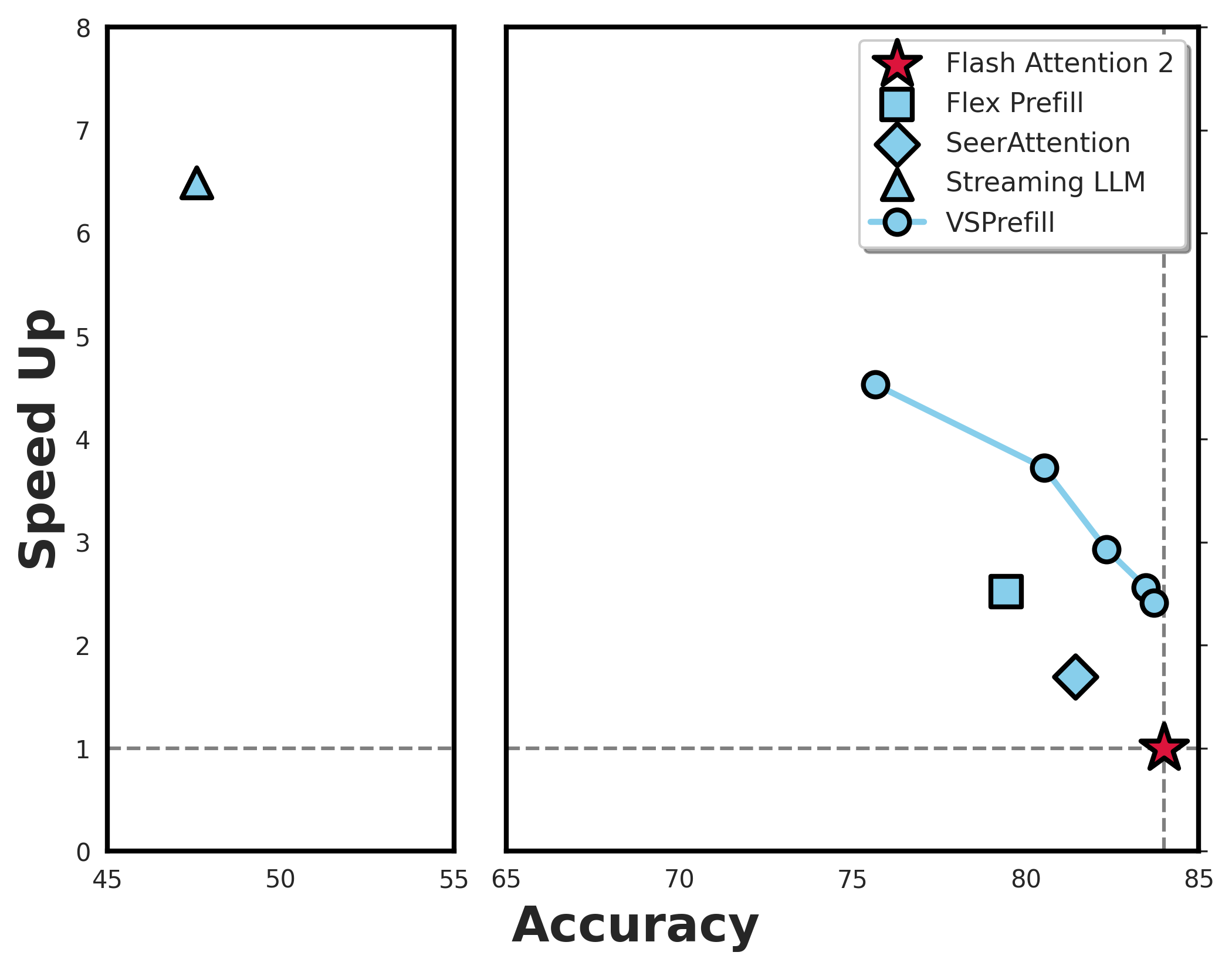}
        \caption{64K Context}
        \label{fig:accuracy_vs_speedup_64k}
    \end{subfigure}
    \hfill
    \begin{subfigure}[b]{0.32\textwidth}
        \centering
        \includegraphics[width=\textwidth]{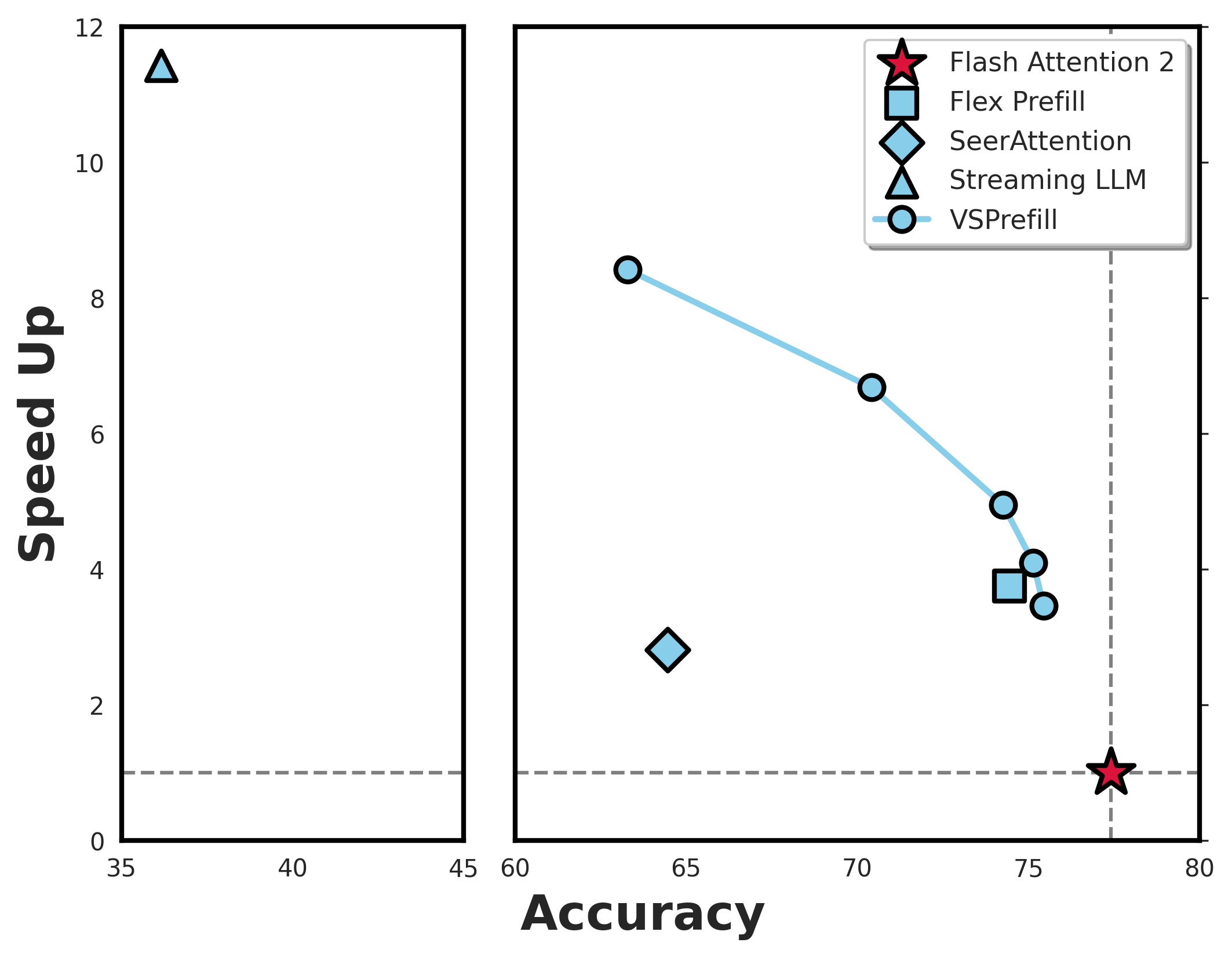}
        \caption{128K Context}
        \label{fig:accuracy_vs_speedup_128k}
    \end{subfigure}
    
    \caption{Accuracy vs. Speedup across different Attention mechanism on context length ranging from 32k to 128k. }
    \label{fig:accuracy_vs_speedup}
\end{figure*}

\section{Experiments}
\label{sec:experiments}
We evaluate VSPrefill against state-of-the-art baselines to assess its performance in long-context scenarios. We first detail our experimental setup, including model architectures, benchmark datasets, and baseline methods. Subsequently, we analyze the performance across diverse tasks, examining both the preservation of accuracy and the acceleration of inference under varying sequence lengths.

\subsection{Experimental Setup}
\label{subsec:exp_setup}
\paragraph{Models}
We evaluate VSPrefill on two widely adopted open-source LLMs: Qwen3-4B-Instruct~\citep{yang2025qwen3} and LLaMA-3.1-8B-Instruct~\citep{touvron2023llama}. These models represent distinct architectural families and training paradigms, thereby providing a robust testbed for the assessment of generalization. All experiments employ the default chat template provided by the official implementation of each model without modification.

\paragraph{Benchmarks}
We adopt two established long-context evaluation suites to assess model capabilities across heterogeneous tasks. LongBench~\citep{bai2024longbench} comprises multi-domain challenges spanning single- and multi-document question answering, summarization, few-shot learning, synthetic reasoning, and code generation. Complementarily, RULER~\citep{ruler2024} extends the needle-in-a-haystack paradigm with customizable sequence lengths and task complexities, introducing tasks of varying intrinsic difficulty alongside novel evaluation dimensions such as multi-hop variable tracing and common word extraction. Together, these benchmarks provide orthogonal perspectives on long-context comprehension and retrieval fidelity.

\paragraph{Baselines}
We compare VSPrefill against four representative attention mechanisms spanning the spectrum from dense computation to sparse approximation. FlashAttention~\citep{dao2022flashattention} serves as the exact attention baseline, employing hardware-aware kernel fusion to optimize memory I/O without compromising mathematical equivalence. StreamingLLM~\citep{xiao2023streamingllm} is a static sparse mechanism that retains initial attention sinks alongside a local sliding window; for evaluation, we configure it with 128 initial tokens and a 2048-token window. FlexPrefill~\citep{lai2025flexprefill} represents a training-free dynamic sparse attention method that estimates global patterns through local sampling and determines sparsity budgets via Jensen--Shannon divergence thresholds. We adopt its recommended hyperparameters of 128 block size, $\gamma = 0.9$, $\tau = 0.1$, and a minimum budget of 1024 tokens. Finally, SeerAttention~\citep{seerattn2024} is a lightweight trainable approach that predicts block-wise sparse patterns through pooled query-key statistics and linear projection. We implement the configuration combining $Q_{\text{avg}}$ and $K_{\text{max-min-avg}}$ pooling strategies as suggested by the authors.

\paragraph{Implementation details }
We distill the VSIndexer module using the LongAlpaca dataset~\citep{longlora}, which provides diverse long-form instruction-following examples with sequence lengths spanning from 2k to 32k tokens. Training utilizes AdamW with a peak learning rate of $1 \times 10^{-3}$, a 500-step linear warmup, and cosine decay. Training comprises 8,000 optimization steps with 8 gradient accumulation steps. VSIndexer employs SiLU activation and a 1024 intermediate dimension. Critically, the backbone is frozen and KV inputs are detached, restricting gradient updates exclusively to VSIndexer parameters. This design enables efficient single-GPU training on an H20 GPU, completing the Qwen3-4B-Instruct adaptation within 6 GPU hours.

To avoid the memory cost of materializing full attention matrices during distillation, we implement a customized fused kernel based on TileLang that extends the FlashAttention algorithm~\citep{dao2022flashattention}. Our kernel performs online aggregation of attention weights along vertical columns and slash diagonals directly within the block-wise computation pipeline, eliminating intermediate storage while preserving numerical accuracy.

\subsection{Main Results}
\label{subsec:main_results}

\paragraph{LongBench}
Table~\ref{tab:longbench-result} summarizes performance across 13 diverse tasks. Baselines exhibit distinct task-specific deficits: StreamingLLM suffers on retrieval-intensive benchmarks due to its restricted local window; FlexPrefill struggles with complex multi-hop reasoning tasks like HotpotQA; and SeerAttention, while accurate, offers limited acceleration. VSPrefill surmounts these limitations, consistently ranking as the runner-up to full attention and occasionally surpassing it. This gain likely stems from the regularization effect induced by sparse computation, which mitigates overfitting to spurious long-range correlations. Overall, VSPrefill demonstrates exceptional fidelity, retaining 98.35\% and 98.13\% of the full attention accuracy on Qwen3-4B-Instruct and LLaMA-3.1-8B-Instruct, respectively.

\paragraph{RULER}
Table~\ref{tab:attention-comparison} details the scalability analysis across sequence lengths ranging from 4k to 128k tokens. Existing methods exhibit distinct limitations: StreamingLLM suffers catastrophic accuracy collapse beyond 32k despite high speed; FlexPrefill degrades at extreme lengths due to accumulated sampling errors; and SeerAttention provides only modest acceleration due to quadratic prediction overhead. In contrast, VSPrefill achieves a superior equilibrium. On Qwen3-4B-Instruct, it delivers a 1.91$\times$ average speedup while limiting accuracy degradation to within 1.1\% relative to the full attention baseline, even at 128k tokens. This robustness consistently generalizes to LLaMA-3.1-8B-Instruct, where our method achieves 1.75$\times$ acceleration with negligible accuracy loss ($<1.5\%$). Collectively, these findings validate VSPrefill as a robust solution that consistently harmonizes inference speed with retrieval fidelity across varying sequence lengths.

\paragraph{Accuracy--speedup trade-off}
Figure~\ref{fig:accuracy_vs_speedup} visualizes the Pareto frontier of accuracy versus speedup for Qwen3-4B-Instruct across three representative sequence lengths. While StreamingLLM occupies the extreme speedup region at the cost of severe quality degradation, and both SeerAttention and FlexPrefill serve as intermediate operating points, VSPrefill establishes a new Pareto-optimal curve that demonstrates exceptional scalability. In the 32k--64k regime, our method achieves lossless acceleration and superior generalization, consistently outperforming all baselines. Critically, this robustness persists into the extreme 128k context, where VSPrefill preserves 98.35\% of full attention accuracy while delivering a 4.95$\times$ speedup. Under aggressive budgets, this acceleration can be further extended to 8.42$\times$. Collectively, these results demonstrate that VSPrefill effectively addresses the performance degradation issues common in prior works without requiring backbone retraining.

These results collectively demonstrate that VSPrefill effectively bridges the gap between static patterns and heavyweight trainable methods. Its lightweight training paradigm yields context-aware sparse patterns with minimal overhead, while the vertical-slash decomposition ensures theoretically grounded linear-complexity sparse pattern prediction. Consequently, VSPrefill delivers state-of-the-art performance across accuracy, speedup, and adaptability dimensions for long-context LLM inference.

\subsection{Ablation Study}

\begin{table}[t]
    \centering
    \caption{Attention Recall (\%) comparison across different sparsity rates.}
    \label{tab:sparsity_ablation}
    \small
    \setlength{\tabcolsep}{2pt} 
    \begin{tabular}{lcccc}
        \toprule
        \multirow{2}{*}{\textbf{Method}} & \multicolumn{4}{c}{\textbf{Sparsity Rate}} \\
        \cmidrule(lr){2-5}
         & \textbf{50\%} & \textbf{90\%} & \textbf{95\%} & \textbf{99\%} \\
        \midrule
        Random & 47.70 & 9.73 & 4.94 & 0.98 \\
        Sampling & 66.33 & 57.36 & 46.57 & 27.78 \\
        \textbf{VSPrefill} & \textbf{97.50} & \textbf{88.48} & \textbf{83.68} & \textbf{72.15} \\
        \bottomrule
    \end{tabular}
\end{table}

\begin{table}[t]
    \centering
    \caption{Ablation study on loss functions (70\% sparsity).}
    \label{tab:loss_ablation}
    \small
    \setlength{\tabcolsep}{6pt} 
    \begin{tabular}{lc}
        \toprule
        \textbf{Loss Function} & \textbf{Recall (\%)} \\
        \midrule
        KL Divergence & \textbf{92.65} \\
        MSE & 85.00 \\
        MSLE & 85.65 \\
        Cosine Similarity & 89.88 \\
        \bottomrule
    \end{tabular}
\end{table}

\begin{table}[t]
    \centering
    \caption{Ablation study on VSIndexer input feature combinations.}
    \label{tab:input_ablation}
    \small
    \setlength{\tabcolsep}{4pt} 
    \begin{tabular}{lcc}
        \toprule
        \textbf{Input Type} & \textbf{Recall (\%)} & \textbf{Loss} \\
        \midrule
        Query (Q) & 91.08 & 5.78 \\
        Key (K) & 92.34 & 3.59 \\
        Value (V) & 91.97 & 4.23 \\
        \midrule
        Query-Key (QK) & 92.26 & 3.61 \\
        \textbf{Key-Value (KV)} & \textbf{92.66} & \textbf{3.46} \\
        \bottomrule
    \end{tabular}
\end{table}

\paragraph{Analysis of Sparsity Strategies}
To evaluate robustness across sparsity levels from 50\% to 99\%, we benchmark VSPrefill against Random selection and Importance Sampling. Random selection exhibits a strict linear decay in recall relative to the retention rate which indicates a fundamental inability to distinguish salient information. While Importance Sampling performs better, it degrades rapidly to 27.78\% at the extreme. In contrast, VSPrefill demonstrates superior resilience by sustaining a Recall of 72.15\% even at 99\% sparsity and validates the efficacy of learned vertical-slash patterns in isolating high-utility information.

\paragraph{Analysis of Loss Functions}
We compare the Attention Recall of models trained using four loss functions including KL divergence, MSE, MSLE, and Cosine Similarity at a fixed 70\% sparsity level. MSE and MSLE prioritize numerical magnitude over distributional nuances and consequently yield suboptimal Recall scores around 85\%. Although Cosine Similarity captures vector orientation to achieve 89.88\%, KL divergence proves optimal. It attains the highest Recall of 92.65\% by explicitly aligning probability distributions to minimize the divergence from ground-truth patterns.

\paragraph{Analysis of Input Combinations}
To determine the optimal configuration for VSIndexer, we examine five feature combinations including single and dual feature sets. We strictly ensure fairness by normalizing the parameter count across experiments and assign a hidden dimension of 2048 to single-feature inputs and 1024 to dual-feature combinations. The results identify the RoPE-encoded Key matrix as the decisive predictor since configurations lacking $K$ significantly underperform. Specifically, the $Q$-only input incurs the highest loss of 5.78. Ultimately, the $KV$ combination attains the lowest Loss of 3.46 and a peak Recall of 92.66\%. This superiority arises because the Key matrix encodes both critical positional cues and dominant semantic information while the Value matrix provides essential supplementary features to perfect the sparse pattern reconstruction.

\section{Conclusion and Future Work}
This paper introduces VSPrefill, a lightweight sparse attention mechanism that exploits the vertical-slash structural pattern inherent in long-context attention to accelerate LLM inference while preserving fidelity. Our approach employs a compact VSIndexer network to predict context-aware vertical and slash patterns per KV group with minimal training overhead, keeping the backbone frozen. Specialized fused kernels facilitate efficient attention aggregation during distillation and high-throughput sparse execution during inference. Experiments across mainstream LLMs and long-context benchmarks demonstrate that VSPrefill maintains near-full attention accuracy while delivering substantial speedups, establishing a new Pareto frontier in the accuracy--efficiency trade-off. Future work includes integrating sparse pattern prediction into pre-training for inherent acceleration without accuracy loss, and extending vertical-slash principles to the decoding stage via adaptive KV cache compression for extremely long-generation tasks.

\bibliographystyle{aaai} 
\bibliography{references}

\clearpage
\onecolumn
\appendix

\section{Appendix}
\subsection{Slash Pattern Analysis}
\label{sec:slash_pattern_analysis}
In this section, we investigate the origin of the slash-shaped attention pattern through a combination of empirical analysis and theoretical derivation. By visualizing attention distributions after averaging queries and keys along different dimensions, we hypothesize that the feature-dimensional distributions of 
$Q$ and $K$ follow a multivariate Gaussian with structured covariance. This hypothesis is validated through feature-wise distribution visualizations in the Qwen3-4B-Instruct model. Based on this statistical characterization, we analytically derive that the expected value of the $QK^\top$ product depends solely on the relative positional offset between tokens. This derivation formally explains the emergence of the slash pattern as a direct consequence of the rotational encoding of RoPE under Gaussian-distributed query and key representations.

\begin{figure}[t]
    \centering
    \begin{subfigure}[b]{0.49\textwidth}
        \centering
        \includegraphics[width=\textwidth]{figures/attention_aggregated/diagonal_heatmap_wo_mean_layer0.png}
        \caption{No averaging}
        \label{fig:attn_agg_slash_wo_mean}
    \end{subfigure}
    \hfill
    \begin{subfigure}[b]{0.49\textwidth}
        \centering
        \includegraphics[width=\textwidth]{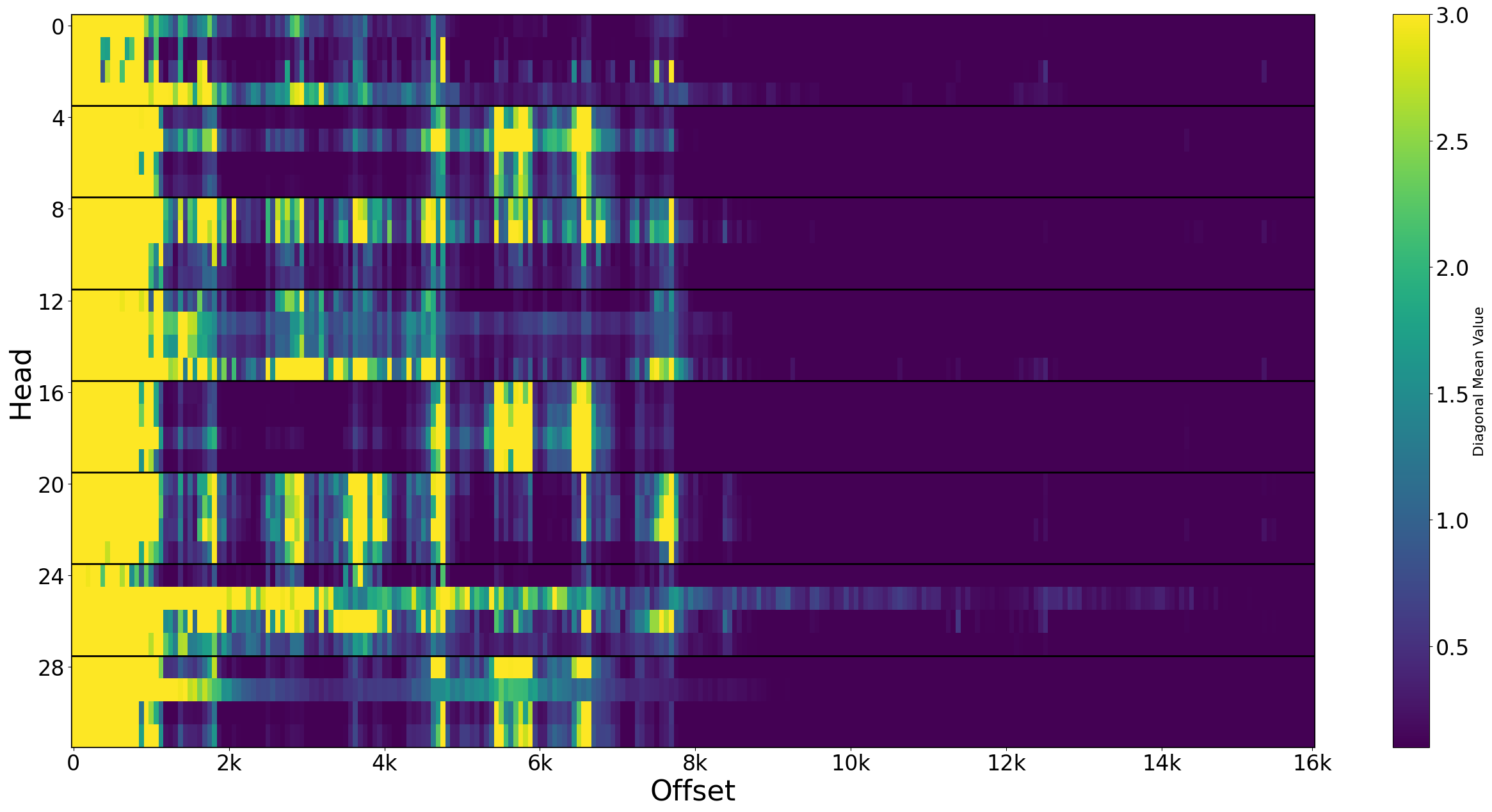}
        \caption{Sequence-wise averaging}
        \label{fig:attn_agg_slash_w_seq_mean}
    \end{subfigure}
    
    \vspace{0.15cm}
    
    \begin{subfigure}[b]{0.49\textwidth}
        \centering
        \includegraphics[width=\textwidth]{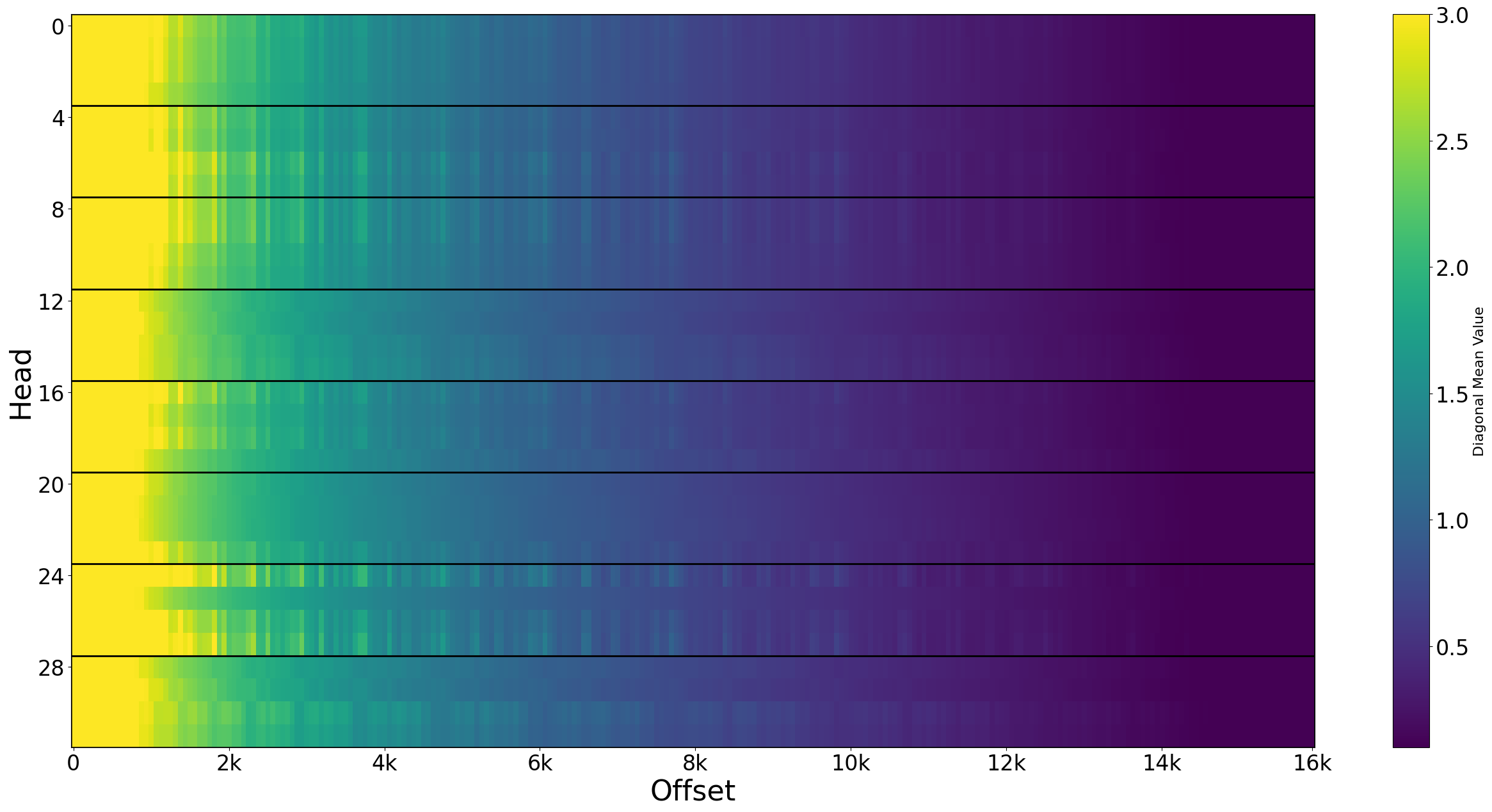}
        \caption{Feature-wise averaging}
        \label{fig:attn_agg_slash_w_dim_mean}
    \end{subfigure}
    \hfill
    \begin{subfigure}[b]{0.49\textwidth}
        \centering
        \includegraphics[width=\textwidth]{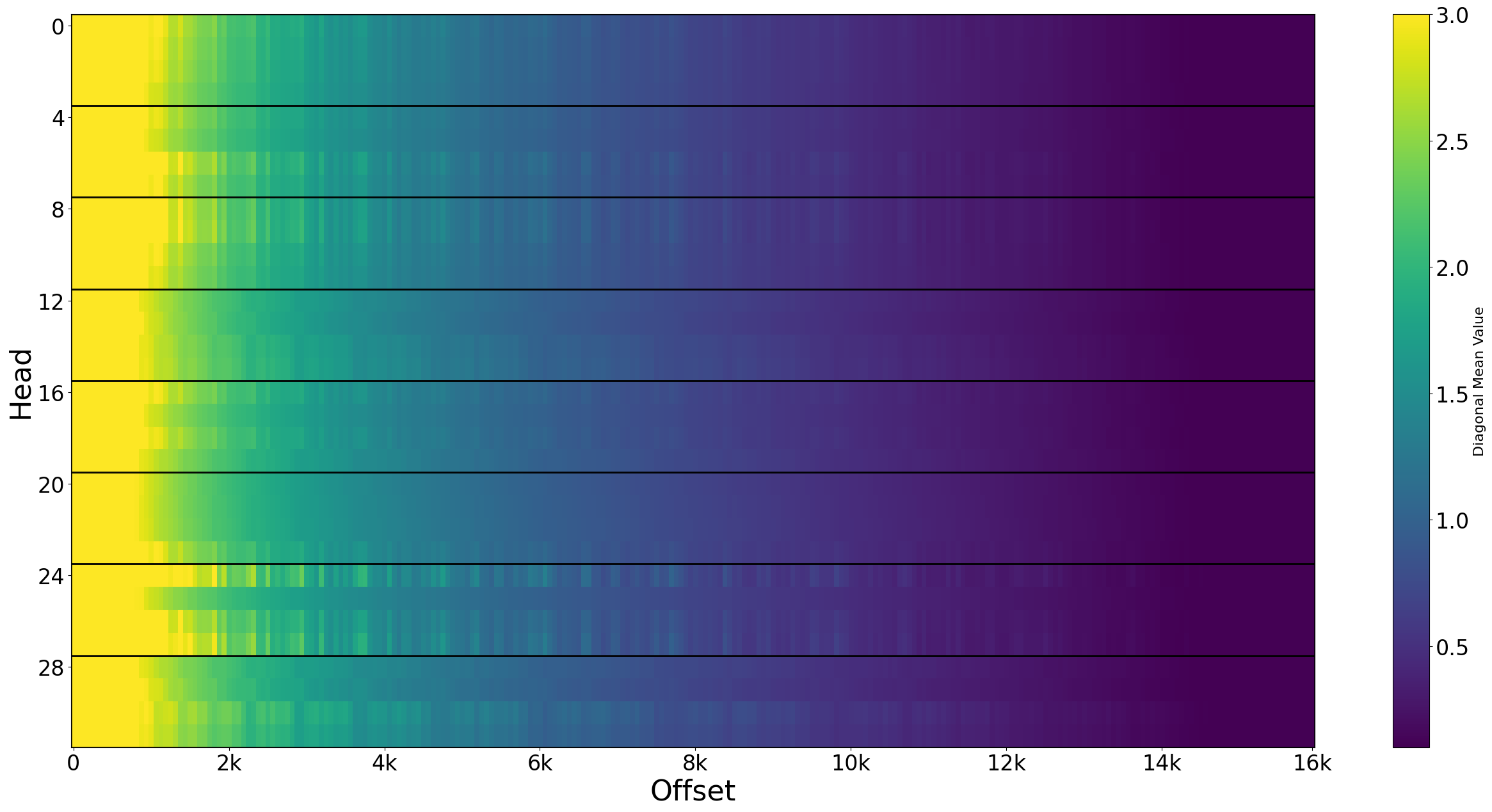}
        \caption{Joint sequence-feature averaging}
        \label{fig:attn_agg_slash_w_seq_dim_mean}
    \end{subfigure}
    
    \caption{Slash-aggregated attention weights under four configurations of query-key averaging before RoPE application: (a) no averaging, (b) averaging along the sequence dimension, (c) averaging along the feature dimension, and (d) averaging along both dimensions. In each case, $Q$ and $K$ are first averaged along the specified dimension(s), then element-wise replicated to restore their original shapes, followed by RoPE encoding.}
    \label{fig:attn_agg_slash_4_config}
\end{figure}

\paragraph{Attention pattern under dimensional averaging}
We visualize slash-pattern attention distributions after averaging queries and keys along the sequence and/or feature dimensions prior to RoPE encoding (Figure \ref{fig:attn_agg_slash_4_config}). The results reveal a clear asymmetry: averaging along the \textbf{sequence dimension} preserves the slash pattern with minimal distortion, whereas averaging along the \textbf{feature dimension}—either alone or jointly with the sequence dimension—significantly degrades the pattern. The latter cases exhibit two characteristic artifacts: (1) shifts in attention peak locations and (2) diffusion of attention mass, causing the distribution to become flatter and spatially blurred. This indicates that the positional encoding of RoPE is highly sensitive to feature-level averaging but robust to sequence-level smoothing. This not only demonstrates a strong association between the slash pattern and the positional encoding of RoPE, but also suggests that the feature-dimensional distribution of $Q$ and $K$ may follow a multivariate Gaussian with a structured covariance that encodes positional relationships.

\paragraph{Gaussian fitting of feature-dimensional activations}
Visualizing the query and key activations of the Qwen3-4B-Instruct model on the LongBench dataset provides strong empirical support for the multivariate Gaussian hypothesis. Histograms of feature-dimensional values in both $Q$ and $K$ are well fitted by Gaussian curves, despite considerable variation in means and variances between dimensions. This consistent fitting behavior is observed in both query and key matrices, reinforcing the hypothesis that their feature-dimensional distributions follow a multivariate Gaussian with structured covariance.

\section*{Theoretical Derivation of Slash Pattern under Multivariate Gaussian Assumption}

\subsection{Notation and Simplifying Assumptions}

\paragraph{Multivariate Gaussian Model}
For arbitrary positions $m, n \in [0, L)$, we model the query and the key vectors as:
\begin{align}
\mathbf{q}_m &\sim \N(\boldsymbol{\mu}^q, \boldsymbol{\Sigma}^q), \quad \mathbf{q}_m \in \R^D \\
\mathbf{k}_n &\sim \N(\boldsymbol{\mu}^k, \boldsymbol{\Sigma}^k), \quad \mathbf{k}_n \in \R^D
\end{align}
where $\boldsymbol{\mu}^q, \boldsymbol{\mu}^k \in \R^D$ are mean vectors and $\boldsymbol{\Sigma}^q, \boldsymbol{\Sigma}^k \in \R^{D \times D}$ are covariance matrices capturing inter-dimensional correlations.

\paragraph{Simplifying assumption} For analytical tractability, we \textit{temporarily assume} $\mathbf{q}_m \perp\!\!\!\perp \mathbf{k}_n$ (statistical independence). This assumption enables a closed-form derivation of the slash pattern mechanism. We emphasize that this is a \textit{theoretical simplification} to isolate the core geometric effect of RoPE.

\paragraph{RoPE Matrix Representation}
The rotary positional embedding matrix $\mathbf{R}(t) \in \R^{D \times D}$ has a block-diagonal structure:
\begin{align}
\mathbf{R}(t) = \bigoplus_{p=0}^{D/2-1}
\begin{bmatrix}
\cos(t\theta_p) & -\sin(t\theta_p) \\
\sin(t\theta_p) & \cos(t\theta_p)
\end{bmatrix},
\quad \theta_p = 10000^{-2p/D}
\end{align}
with properties $\mathbf{R}(m)\mathbf{R}(n)^\top = \mathbf{R}(m-n)$ and $\mathbf{R}(t)^\top \mathbf{R}(t) = \mathbf{I}$.

\paragraph{Derivation of Expectation $\E[P_{m,n}]$}

Under the independence assumption, the attention score $P_{m,n} = \mathbf{q}_m^\top \mathbf{R}(m-n) \mathbf{k}_n$ has the following expectation:
\begin{align}
\E[P_{m,n}] 
&= \E[\mathbf{q}_m]^\top \mathbf{R}(m-n) \E[\mathbf{k}_n] 
= \boldsymbol{\mu}^{q\top} \mathbf{R}(m-n) \boldsymbol{\mu}^k
\end{align}
Expanding in RoPE rotation planes $(2p, 2p+1)$ using complex representation:

\begin{align}
\E[P_{m,n}]
&= \sum_{p=0}^{D/2-1} \Re\Big( \mu^q_{(p)} \cdot \mu^{k*}_{(p)} \cdot e^{i(m-n)\theta_p} \Big) \\
&= \sum_{p=0}^{D/2-1} \Re\Big( (\mu^q_{2p} + i\mu^q_{2p+1})(\mu^k_{2p} - i\mu^k_{2p+1}) e^{i(m-n)\theta_p} \Big) \\
&= \sum_{p=0}^{D/2-1} \Re\Big( \underbrace{(\mu^q_{2p}\mu^k_{2p} + \mu^q_{2p+1}\mu^k_{2p+1})}_{a_p} + i\underbrace{(-\mu^q_{2p}\mu^k_{2p+1} + \mu^q_{2p+1}\mu^k_{2p})}_{b_p} \Big) e^{i(m-n)\theta_p} \\
&= \sum_{p=0}^{D/2-1} \big( a_p \cos((m-n)\theta_p) - b_p \sin((m-n)\theta_p) \big) \\
&= \sum_{p=0}^{D/2-1} r_p \cos\big( \alpha_p + (m-n)\theta_p \big)
\end{align}

where:
\begin{align}
r_p &= \sqrt{a_p^2 + b_p^2}, \\
\alpha_p &= \arctan(-\frac{b_p}{a_p}), \\
a_p &= \mu^q_{2p}\mu^k_{2p} + \mu^q_{2p+1}\mu^k_{2p+1}, \\
b_p &= -\mu^q_{2p}\mu^k_{2p+1} + \mu^q_{2p+1}\mu^k_{2p}
\end{align}

Under the assumption that the query and key vectors follow multivariate Gaussian distributions with structured covariance, the expectation $\mathbb{E}[P_{m,n}]$ of the attention score $P_{m,n}=Q_m K_n^\top$ depends \textit{solely} on the relative positional offset $(m-n)$. This positional invariance induces a periodic structure aligned with the diagonal, which constitutes the theoretical origin of the slash pattern. In practice, however, the input $Q$ and $K$ deviate from ideal multivariate Gaussians, are not perfectly orthogonal, and undergo subsequent softmax normalization. These factors prevent the expectation-based conclusion from being directly applicable to the actual computation of attention. Nevertheless, this theoretical framework successfully explains the emergence of the slash pattern observed in causal LLM inference.

\begin{figure*}[t]
    \centering
    \includegraphics[width=\textwidth]{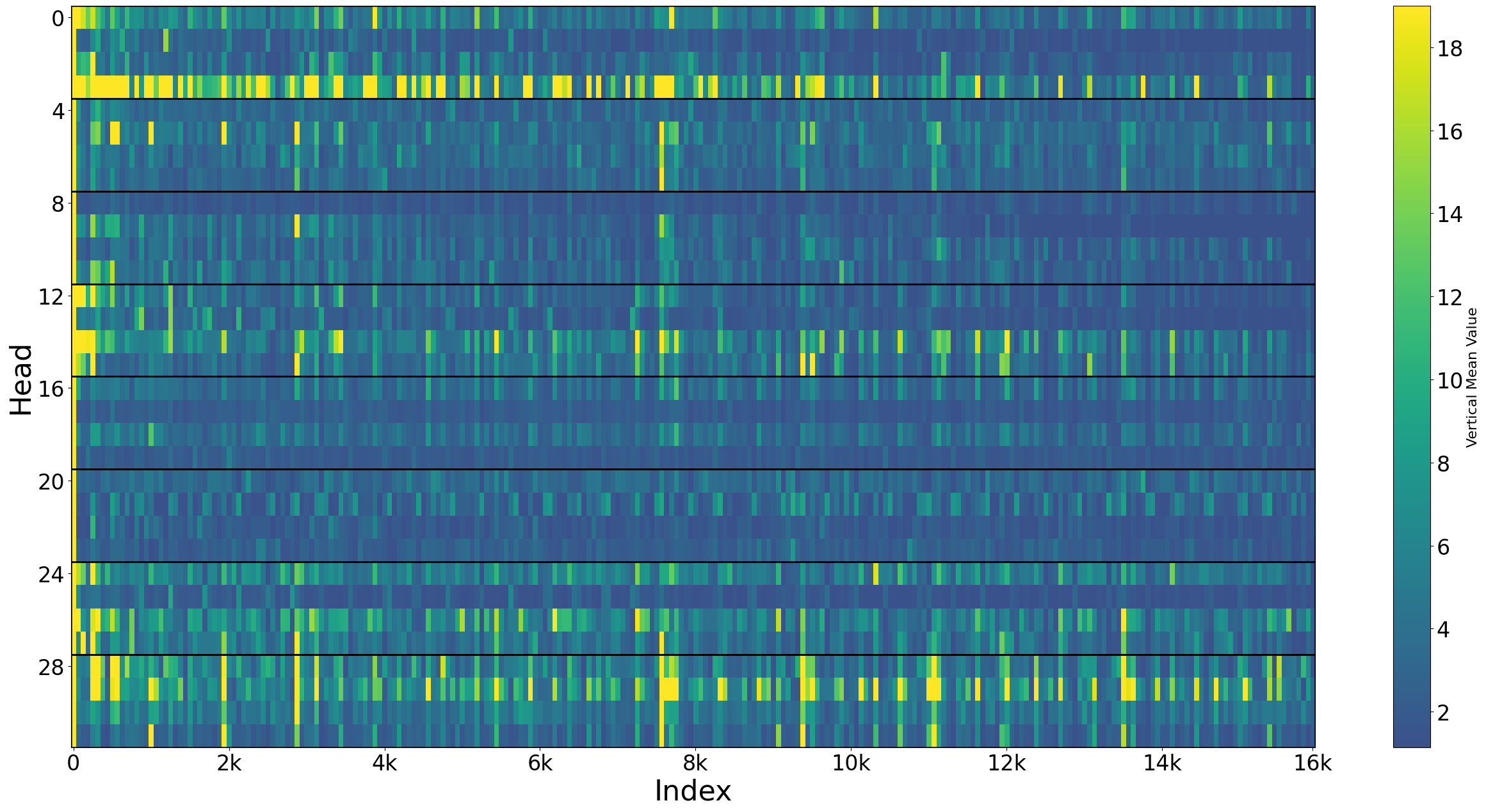}
    
    \caption{Attention weights aggregated along the vertical direction across different heads on Qwen3-4B-Instruct using the LongBench dataset. }
    \label{fig:attention_aggregated_vertical_wo_mean}
\end{figure*}

\begin{figure}[t]
    \centering
    \begin{subfigure}[b]{0.49\textwidth}
        \centering
        \includegraphics[width=\textwidth]{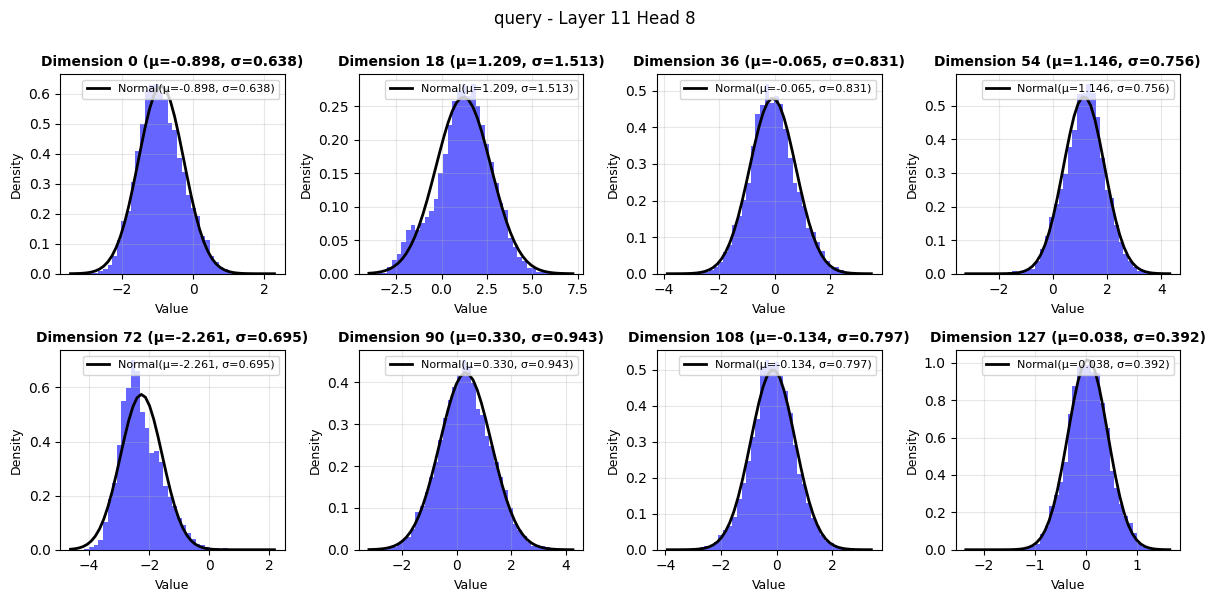}
        \caption{Query}
        \label{fig:dim_hist_q}
    \end{subfigure}
    \hfill
    \begin{subfigure}[b]{0.49\textwidth}
        \centering
        \includegraphics[width=\textwidth]{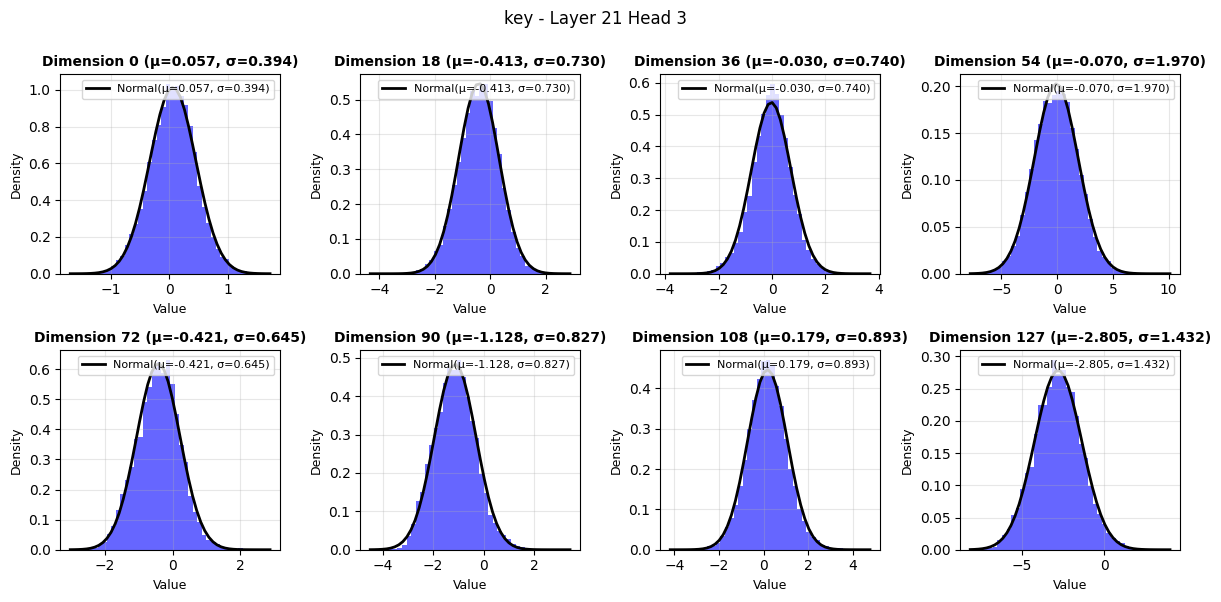}
        \caption{Key}
        \label{fig:dim_hist_k}
    \end{subfigure}
    
    \vspace{0.3cm}
    
    \caption{Dimension-wise histogram statistics of Query and Key vectors with Gaussian fits. The distributions display dimension-dependent means and variances, revealing heterogeneous statistical properties of attention projections.}
    \label{fig:dim_hist}
\end{figure}

\end{document}